\documentclass[10pt,twocolumn,letterpaper]{article}

\usepackage[pagenumbers]{iccv} 

\usepackage{xspace}
\usepackage{colortbl}
\newcommand\ours{CDAL\xspace}
\newcommand\sectionthreetwo{Counterfactual Feature Isolation\xspace}
\newcommand\sectionthreethree{Causal Attention Augmentation\xspace}
 \newcounter{alphasect}

 \renewcommand\thesection{%
 \ifnum\value{alphasect}=1%
A
 \else
\ifnum\value{alphasect}=2%
B
\else
\ifnum\value{alphasect}=3%
C
\else
\ifnum\value{alphasect}=4%
D
\else
 \arabic{section}
 \fi\fi\fi\fi}%

 \newcolumntype{H}{>{\setbox0=\hbox\bgroup}c<{\egroup}@{}}
\definecolor{iccvblue}{rgb}{0.21,0.49,0.74}
\usepackage[pagebackref,breaklinks,colorlinks,allcolors=iccvblue]{hyperref}

\usepackage{multirow}
\usepackage{arydshln}
\def\paperID{183} 
\def\confName{ICCV}
\def\confYear{2025}

\title{Learning Counterfactually Decoupled Attention for Open-World \\Model Attribution}

\author{
Yu Zheng\thanks{Equal contribution.}\;\; Boyang Gong\footnotemark[1] \;\; Fanye Kong \;\; Yueqi Duan \;\; Bingyao Yu \;\; Wenzhao Zheng \;\; Lei Chen \\
Jiwen Lu \;\; Jie Zhou\\
Tsinghua University, China\\
}

\begin{document}
\maketitle
\begin{abstract}
In this paper, we propose a Counterfactually Decoupled Attention Learning (\ours) method for open-world model attribution. 
Existing methods rely on handcrafted design of region partitioning or feature space, which could be confounded by the spurious statistical correlations and struggle with novel attacks in open-world scenarios. 
To address this, \ours explicitly models the causal relationships between the attentional visual traces and source model attribution, and counterfactually decouples the discriminative model-specific artifacts from confounding source biases for comparison. 
In this way, the resulting causal effect provides a quantification on the quality of learned attention maps, thus encouraging the network to capture essential generation patterns that generalize to unseen source models by maximizing the effect. 
Extensive experiments on existing open-world model attribution benchmarks show that with minimal computational overhead, our method consistently improves state-of-the-art models by large margins, particularly for unseen novel attacks. Source code: \url{https://github.com/yzheng97/CDAL}. 

\end{abstract}    
\section{Introduction}
\label{sec:intro}
The rapid development of visual generative models~\cite{kingma2013auto,sohn2015learning,van2017neural,goodfellow2014generative,karras2019style,huang2020pfa,song2020denoising,esser2021taming,dhariwal2021diffusion,rombach2022high,tian2024visual} has raised increasing concerns about their potential misuse for personal reputation damage and public misinformation. 
Although deepfake detectors can effectively distinguish real from synthetic content~\cite{li2020face,zhao2021multi,liu2022detecting,ojha2023towards,tan2024rethinking,yan2024df40}, such capability alone does not address the need to identify source generative models for further legal measures, referred to as model attribution. 

\begin{figure}[t]
    \centering
    \includegraphics[width=\linewidth]{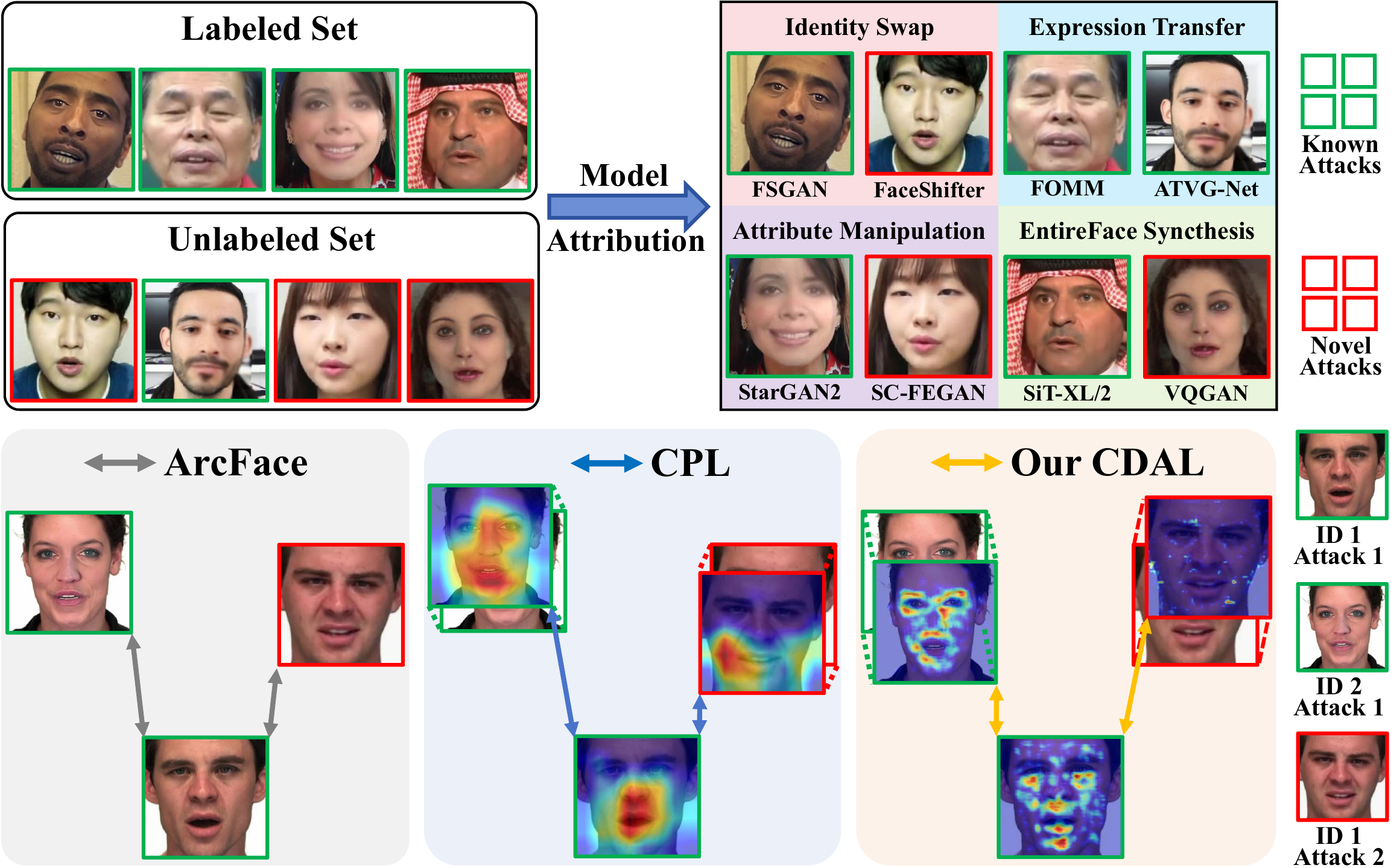}
    \vspace{-7mm}
    \caption{\textbf{Upper: }Problem Setup. 
    \textbf{Bottom: }Exemplified sample in feature space.   
    \textbf{Bottom left:} ArcFace~\cite{deng2019arcface} distances show that forgery images of the same source identity are clustered together regardless of their attacking models. \textbf{Bottom middle:} Feature differences learned by existing methods~\cite{sun2023contrastive} are still influenced by source bias (i.e., identity) from the novel attack rather than model-specific artifacts. \textbf{Bottom right:} Our approach effectively isolates model-specific artifacts from source content bias, thus enabling accurate attribution even for unseen generative models. 
    }
    \vspace{-4mm}
    \label{fig:motivation_teaser}
\end{figure}

Pioneering works on model attribution actively root fingerprints during training~\cite{yu2019attributing,yu2021responsible,yu2021artificial,kim2021decentralized} or parse the given AI-generated images~\cite{marra2019gans,yu2019attributing,guarnera2022exploitation,yang2022deepfake,bui2022repmix,wang2023did,liu2024model,li2024handcrafted}. 
However, their closed-set assumption becomes impractical as new generative models emerge rapidly and challenge their generalizability. 
To this end, researchers have built benchmarks on open-world model attribution and accordingly devised techniques aimed to attribute known attacks to their source models while identifying unseen novel attacks in real world scenarios (Figure~\ref{fig:motivation_teaser} upper). 
For example, \cite{sun2023contrastive} designs a Contrastive Pseudo Learning framework for open-world deepfake attribution, which is later extended with multi-scale learning and frequency information~\cite{sun2024rethinking}. 
\cite{yang2023progressive} simulates a feature space spanning mechanism specifically designed for GANs. 
To exploit the subtle artifacts from various generative models, these approaches design handcrafted region partitions to search for useful local patterns~\cite{sun2023contrastive,sun2024rethinking}, or simulate the progressive feature space spanning via manually modifying the model fingerprints~\cite{yang2023progressive}.

However, as exemplified in Figure~\ref{fig:motivation_teaser}, apart from their inherent model-specific artifacts, generative models preserve semantic biases from source images during forgery processes. 
This poses a critical attribution challenge where the semantic content dominates over model-specific traces. 
The ArcFace~\cite{deng2019arcface} embeddings (Figure~\ref{fig:motivation_teaser} bottom left) serve as evidence of this challenge where forgery images are clustered by identity rather than by generative source in feature space. 
Even for state-of-the-art methods like CPL~\cite{sun2023contrastive} (Figure~\ref{fig:motivation_teaser} bottom middle), 
despite their sophisticated design of handcrafted pixel partition or feature space, they remain confounded by source bias content. 
Their resulting attention maps fail to consistently highlight model-specific forgery patterns, thus leading to poor generalization when confronting novel unseen attacks where the previously learned spurious correlations no longer apply. 

In this paper, we propose a plug-and-play Counterfactually Decoupled Attention Learning (\ours) method for open-world model attribution. 
Different from previous methods confounded by the spurious correlations, \ours aims to explicitly model the essential causal relationships between visual forgery traces and attributed source models that generalize well in open-world scenarios. 
Specifically, taking inspiration from counterfactual intervention, we isolate model-specific artifacts from source content biases by extracting factual and counterfactual attentions from the input feature representation. 
Given the diversified and uncertain nature of forgery traces in open-world scenarios, we further propose the Causal Attention Augmentation to achieve broader spatial coverage for these decoupled attentions while maintaining the constructed causal consistency. 
The quality of the learned attention maps can be quantified through the causal effect, i.e., the difference between the attribution results predicted by factual and counterfactual attentions. 
By maximizing this effect, \ours provides a supervision signal to encourage the attribution network to attend to discriminative generation patterns open-world model attribution (Figure~\ref{fig:motivation_teaser} bottom right), while resisting misleading source biases. 
With negligible computational overhead, our method can be incorporated readily into existing methods where \ours significantly improves baseline performances across different OWMA benchmarks, including OW-DFA~\cite{sun2023contrastive} for deepfake attribution, and OSMA~\cite{yang2023progressive} for GAN attribution and discovery. 
\section{Related Works}
\label{sec:related}

\textbf{Model Attribution: }
Model attribution aims to identify the source generative model of visual content. 
Active methods~\cite{yu2019attributing,yu2021responsible,yu2021artificial,kim2021decentralized} inject specifically-designed fingerprints during the training phase of the generative model. 
However, they cannot handle the ``black-box'' scenario without access to the source model and its training pipeline. 
In contrast, passive attribution methods~\cite{wang2023did,bui2022repmix,liu2024model,li2024handcrafted} focus on identifying the source model solely from its generated outputs. 
For example, \cite{marra2019gans,yu2019attributing,guarnera2022exploitation,yang2022deepfake} attribute GAN architectures through parsing their inherent model fingerprints within the generated images. 
Our \ours belongs to the model-agnostic passive technique, and is applicable to various scenarios including deepfake model attribution, GAN attribution an GAN discovery.

\noindent\textbf{Open-world Recognition: }
Open-world Recognition aims at identifying known categories and simultaneously recognizing novel classes unseen during training, which has been explored in visual classification~\cite{guo2022robust,cao2022openworld,rizve2022openldn}, semantic segmentation~\cite{sodano2024open} and object detection~\cite{joseph2021towards}, etc. 
As new generative models emerge rapidly, researches have paid attention to constructing benchmarks to simulate the open-world model attribution scenarios and developing the corresponding solutions~\cite{girish2021towards,yang2023progressive,sun2023contrastive,sun2024rethinking}. 
However, these approaches primarily rely on statistical correlations through handcrafted designs of region partitioning~\cite{sun2023contrastive,sun2024rethinking} or feature space~\cite{yang2023progressive}, which limits their effectiveness when the learned correlations no longer hold for diverse unseen models. 
Our \ours instead focuses on modeling causal relationships between visual forgery traces and source models, which enables more robust generalization to previously unseen forgery techniques. 

\noindent\textbf{Causal Reasoning in Computer Vision: }
There has been a number works leveraging causal reasoning~\cite{neuberg2003causality,pearl2018book} to benefit various sub-tasks in computer vision, including classification~\cite{lopez2017discovering}, captioning~\cite{wang2020visual}, re-identification~\cite{rao2021counterfactual,li2022counterfactual}, etc. 
To isolate the direct causal effect of interest critical to the target task, they perform counterfactual interventions through the $do$-operator, which is typically implemented by randomized attention maps to cut off the causal path between confounding variables and outcomes~\cite{wang2020visual,rao2021counterfactual,li2022counterfactual,wang2022counterfactual}. 
We are inspired by this intervetion operation and propose to decouple the factual and counterfactual attentions in the context of open-world model attribution. 
\section{Approach}
\label{sec:approach}
\setlength{\abovedisplayskip}{3.5pt}
\setlength{\belowdisplayskip}{3.5pt}
In this section, we present our proposed Counterfactually Decoupled Attention Learning (\ours) framework. 
We firstly introduce the addressed task from a causal perspective. 
Then we outline our \ours with our core insights of counterfactually decoupled attention learning followed by the technical details of counterfactual feature isolation and causal attention augmentation. 

\begin{figure*}[ht]
    \centering
    \includegraphics[width=.93\textwidth]{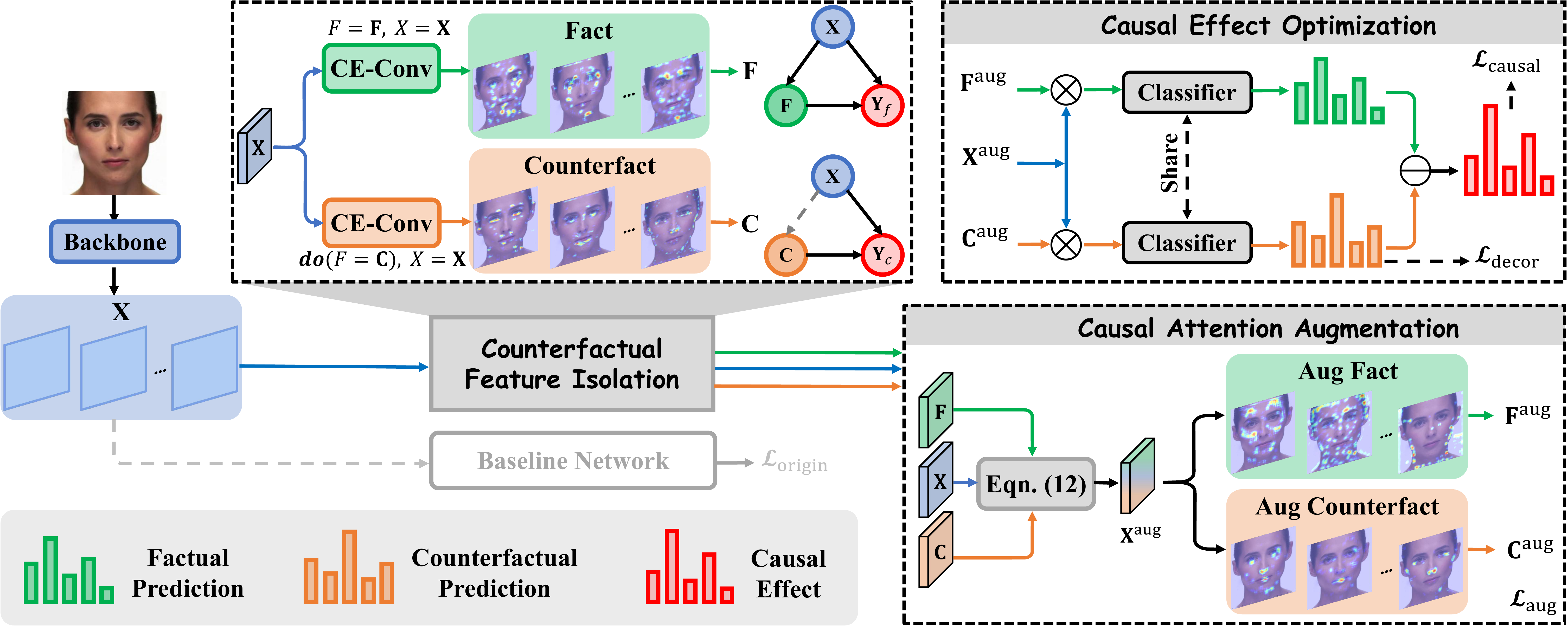}
    \vspace{-3mm}
    \caption{Overview of our proposed \ours, which can be readily incorporated into existing baseline networks. \textbf{Upper right: }\ours fundamentally maximizes the causal effect, i.e., the difference between the predictions from factual and counterfactual attentions, to encourage the network to learn more effective visual attention for model attribution and reduce the effects of biased training data. 
    \textbf{Upper left: }The factual and counterfactual attentions are generated by Counterfactual Feature Isolation, which simulates counterfactual intervention and employs parallel Causal Expert (CE) convolutions. 
    \textbf{Bottom right: }We complement the extracted attention maps with  broader spatial coverage through Causal Attention Augmentation while maintaining the constructed causal consistency. 
    }
    \vspace{-5mm}
    \label{fig:main_framework}
\end{figure*}

\subsection{Causal Perspective on the Addressed Problem}
\textbf{Problem Setup: }Open-world model attribution~\cite{yang2023progressive,sun2023contrastive,sun2024rethinking} aims to identify the source generative model of synthetic images, including those from previously unseen models. 
Given known generative models $\mathbb{G}_K=\{\mathcal{G}_1, \mathcal{G}_2, \cdots, \mathcal{G}_M\}$ and their generated images $\mathcal{X}_K = \{x_i | x_i \sim \mathcal{G}_j, \mathcal{G}_j \in \mathbb{G}_K\}$, we learn a function $f$ that maps: 
\begin{equation}
\label{eq:owma_definition}
f:\mathcal{X}\rightarrow \mathbb{G}_K\cup \{\text{unknown}\}
\end{equation}
When $f(x) = \text{unknown}$, the image is determined to be generated by some unknown model $\mathcal{G}_u \in \mathbb{G}_U$, where $\mathbb{G}_U$ represents new generative models not available during training. 
As aforementioned, a key challenge lies in the problem that two types of features are blended in synthetic images: 
\begin{itemize}
    \item Model-specific artifacts that \textit{causally relate to} source models. 
    \item Source bias features from original content with \textit{no causal link to} generation methods. 
\end{itemize}
Existing approaches typically rely on handcrafted region partitioning~\cite{sun2023contrastive,sun2024rethinking} or feature spanning~\cite{yang2023progressive} strategies, which could be confounded by the learned statistical correlations. 
As they fail to isolate the true causal relationships between visual artifacts and their generating processes, these spurious correlations might break down when confronted with previously unseen generative models. 

\noindent\textbf{Structural Causal Model for Attribution: }
To address this limitation, we reformulate the problem with a Structural Causal Model (SCM): $G = \{N, E\}$, where nodes $N$ and edges $E$ represent variables and their causal dependencies: 
\begin{equation}
\label{eq:causal_setup}
X \rightarrow A \rightarrow Y \text{ and } X \rightarrow Y. 
\end{equation}
Here, $X$, $A$, $Y$  denotes the input image, feature maps used for attribution (specifically, the attention maps), and source model prediction. 
The attention map $\mathbf{A}_i \in \mathbb{R}_+^{H \times W}$ highlights regions in the image that are important for attribution decisions, where $H$ and $W$ are the height and width of the attention map. These attention maps are used to selectively focus on discriminative regions by weighing feature maps through element-wise multiplication: 
\begin{equation}
\mathbf{h}_i = \varphi(X * \mathbf{A}_i) = \frac{1}{HW}\sum_{h=1}^{H}\sum_{w=1}^{W}X^{h,w}\mathbf{A}_i^{h,w},
\end{equation}
where $*$ and $\varphi$ denotes Hadamard product and a global average pooling operation that aggregates the weighted features. The attention-weighted representations from different regions are then combined to form the final representation for attribution prediction. 

\noindent\textbf{Our \ours: }
Built upon the structural causal model above, our \ours framework aims to capture the essential causal relationships between visual forgery traces and the attributed source models, whose overall framework is illustrated in Figure~\ref{fig:main_framework}. 
The basic idea is to quantify the quality of the learned feature maps by comparing the effects of factual attention $\mathbf{F}$ (i.e., the learned model-specific attention) and counterfactual attention $\mathbf{C}$ (i.e., the attention on source bias). 
This difference (\textbf{upper-right} in Figure~\ref{fig:main_framework}), also known as causal effect~\cite{neuberg2003causality,pearl2018book}, is maximized to encourage the network to learn more effective visual attention for model attribution and reduce the effects of biased training data. 

Specifically, given the input feature $\mathbf{X}$, we are inspired by counterfactual intervention $do(A = \bar{A})$~\cite{pearl2018book} to isolate model-specific artifacts from source content bias (\textbf{upper left} in Figure~\ref{fig:main_framework})
, where we extract the factual attentions $\mathbf{F}$ and counterfactual attentions $\mathbf{C}$ from $\mathbf{X}$: 
\begin{equation}
\begin{aligned}
\label{eq:attentions_extraction}
    \mathbf{F} &= \mathcal{F}(\mathbf{X}) = \{\mathbf{F}_1, \dots, \mathbf{F}_{M}\}, \\
    \mathbf{C} &= \mathcal{C}(\mathbf{X}) = \{\mathbf{C}_1, \dots, \mathbf{C}_{M}\},
\end{aligned}
\end{equation}
where $\mathcal{F}(\cdot)$ and $\mathcal{C}(\cdot)$ extract factual and counterfactual attentions \textbf{(introduced in Section~\ref{sec:counterfact_attn_extraction})} respectively. 
$M$ is the number of attention maps.

After obtaining these attention maps, given the uncertain and diverse nature of subtle forgery traces in open world, we complement them with broader spatial coverage while maintaining causal consistency (\textbf{bottom-right} in Figure~\ref{fig:main_framework}). 
This is achieved through Causal Attention Augmentation \textbf{(introduced in Section~\ref{sec:attention_aug})} which outputs the augmented features $\mathbf{X}^{\text{aug}}$ and corresponding attention maps $\mathbf{F}^{\text{aug}}$, $\mathbf{C}^{\text{aug}}$. 
A shared attribution classifier $\delta(\cdot)$ is used to map them into factual and counterfactual predictions ($\mathbf{Y}_f$ and $\mathbf{Y}_c$): 
\begin{equation}
\begin{aligned}
\mathbf{Y}_f \hspace{-0.3mm}&=\hspace{-0.3mm} \mathbf{Y}(F\hspace{-1mm}=\hspace{-1mm}\mathbf{F}, X\hspace{-1mm}=\hspace{-1mm}\mathbf{X}) 
\triangleq \delta(\sum\limits^{M}_{i=1}\mathbf{X}^{\text{aug}}*\mathbf{F}^{\text{aug}}_i),\\
\mathbf{Y}_c \hspace{-0.3mm}&=\hspace{-0.3mm} \mathbf{Y}(do(F\hspace{-1mm}=\hspace{-1mm}\mathbf{C}), X\hspace{-1mm}=\hspace{-1mm}\mathbf{X}) 
\triangleq \delta(\sum\limits^{M}_{i=1}\mathbf{X}^{\text{aug}}*\mathbf{C}^{\text{aug}}_i). 
\vspace{-2mm}
\end{aligned}
\end{equation}
The difference $\mathbf{Y}_{\text{effect}} = \mathbf{Y}_f - \mathbf{Y}_c$, inspired by causal effect analysis~\cite{neuberg2003causality,pearl2018book}, quantifies the quality of learned feature representations by measuring how much the model-specific artifacts (factual) outperform the bias-prone features (counterfactual) in attribution predictions. 
We leverage this causal effect as the optimization target by employing a cross-entropy (CE) loss function $\mathcal{L}_{\text{causal}}$:

\begin{equation}
\label{eq:causal_cls_loss}
\mathcal{L}_{\text{causal}} = \text{CE}(\mathbf{Y}_{\text{effect}}, y),
\end{equation}
where $y$ is the ground-truth attribution label. 
This optimization target creates opposing learning objectives where factual and counterfactual attentions are driven toward discriminative model-specific traces and source-dependent biases respectively. 
By maximizing their performance gap, we strengthen the causal connection between the attentional features and source model attribution.

\subsection{\sectionthreetwo}
\label{sec:counterfact_attn_extraction}
To extract the counterfactual attention maps in Eqn.~(\ref{eq:attentions_extraction}), a common practice is to replace the original attention with the ones with randomized fixed weights~\cite{wang2020visual,wang2022counterfactual,li2022counterfactual}. 
However, such a predefined distribution on counterfactual attention weights, along with the handcrafted design~\cite{sun2023contrastive,yang2023progressive,sun2024rethinking} in search of factual forgery traces, still typically rely on statistical correlations that inadequately capture diverse artifact characteristics in unseen attacks (see Table~\ref{tab:counterfact}). 
To address this limitation, we employ Causal Expert Convolution that is guided by the aforementioned opposite learning objectives and actively learns to extract counterfactual patterns without predefined assumptions, rather than simply cut off the causal effect between input and counterfactual attention.

\noindent\textbf{Causal Expert Convolution: }
Unlike static convolutions that indiscriminatively mix all feature correlations, our employ Causal Expert (CE) convolutions to explicitly model the causal effect through which model-specific artifacts influence the attribution prediction. 
This is achieved by dynamically constructing a convolution kernel $W^{\prime}$ that weights $N_{exp}$ different expert kernels based on their causal relevance to the attribution task: 
\begin{equation}
    W^{\prime} =\sum_{i=1}^{N_{exp}} \alpha_{i} W_{i}, \;\;\alpha = \sigma(\text{MLP}(\text{Pool}(\mathbf{X})))\in\mathbb{R}^{N_{exp}}
\end{equation}
where $\alpha$ denotes the denotes the estimated causal contribution factors and $W_{i}$ is the $i$-th expert kernel. 

\noindent\textbf{Causal Attention Generation: } 
The CE convolution is used to extract both factual ($F$) and counterfactual ($C$) attention maps in Eqn.~(\ref{eq:attentions_extraction}). 
Below we describe the process for generating $F$, which consists of two complementary extraction paths. 
First, a $1\times1$ CE convolution operator consumes $\mathbf{X}$ to obtain $\mathbf{X}_{\text{cross}}$, which preserves global relationships across channels. 
$\mathbf{X}_{\text{cross}}$ is then fed into a depth-wise separable~\cite{chollet2017xception} CE convolution, producing $\mathbf{X}_{\text{depth}}$ with its channel-specific patterns. 
This design helps isolate channel-specific patterns~\cite{han2020ghostnet} that may contain distinctive model artifacts, which might otherwise be diluted in standard convolutions. 
The final factual attention map combines these complementary features: 
\begin{equation}
\mathbf{F} = \text{Concat}[\mathbf{X}_{\text{cross}}, \mathbf{X}_{\text{depth}}]. 
\end{equation} 
Similarly, the counterfactual maps $\mathbf{C}$ are generated by this two-stage process with another set of network parameters. 

Since forgery methods preserve semantic source bias while introducing model-specific artifacts, we enhance the separation between discriminative features and source biases through a decorrelation loss: 
\begin{equation}
    \mathcal{L}_{\text{decor}} = \text{CE}(\mathbf{Y}_{c},\mathbf{Y}_{c}) = -\sum_{c \in \mathbf{C}} \mathbf{Y}_{c} \log \mathbf{Y}_{c}. 
\end{equation}
By maximizing the entropy in counterfactual predictions, $\mathcal{L}_{\text{decor}}$ pushes $\mathbf{Y}_{c}$ towards a uniform distribution across attributed classes, and forces counterfactual attention to focus on non-causal source content that lacks discriminative value for model attribution. 

\begin{figure}[t]
    \centering
    \includegraphics[width=.88\linewidth]{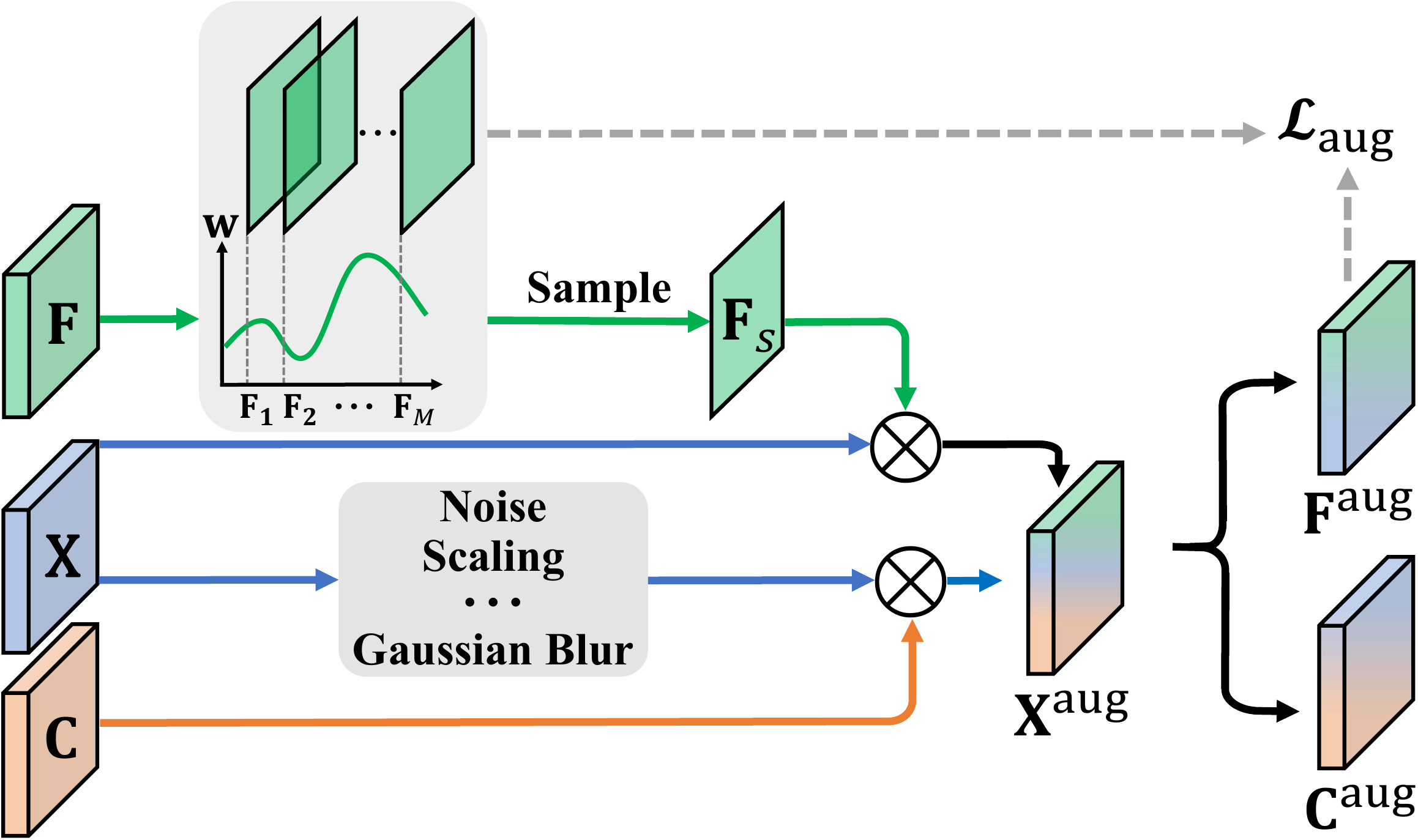}
    \vspace{-3mm}
    \caption{Illustration of Casual Attention Augmentation.}
    \vspace{-5mm}
    \label{fig:caa}
\end{figure}

\subsection{\sectionthreethree}
\label{sec:attention_aug}
While our attention learning aims to help identify potential forgery traces, two critical challenges remain in open-world scenarios: (1) the uncertainty in unseen source features (new faces, unexpected noises) that can weaken decoupling efficacy, and (2) the inherent diversity of forgery traces in full-face synthesis which can appear across multiple locations. 
To address these challenges, we propose a Causal Attention Augmentation operation (shown in Figure~\ref{fig:caa}) that expands attention coverage while maintaining causal relationships. 
It firstly generates diversified feature representations through standard augmentation techniques: 
\begin{equation}
\label{eq:naive_aug}
 \mathbf{X}^{\text{aug}} = \mathcal{A}_1(\mathcal{A}_2(\cdots \mathcal{A}_N(\mathbf{X)))}, 
\end{equation}
where ${\mathcal{A}_1,\cdots,\mathcal{A}_N}$ are a series of operations such as noise adding or Gaussian blurring (see Appendix for details). 

However, indiscriminative application of these augmentations risks diluting model-specific traces and corrupting established causal relationships. 
We therefore employ a targeted augmentation strategy to preserve forgery-relevant regions while diversifying source-specific features only. 
To explore diverse forgery traces while maintaining causal consistency, we probabilistically select an influential factual attention map indexed by $s$: 
\begin{equation}
\label{eq:select_index}
 s = \underset{i \in {1,2,...,M}}{\operatorname{arg,sample}}(i \mid p(i) = \mathbf{w}_i)
\end{equation} 
where $\mathbf{w}\in \mathbb{R}^M$ is the normalized energy distribution of factual attention channels (see Appendix for derivation). 

The corresponding sampled attention map $\mathbf{F}_s$ is then used for selective augmentation: 
\begin{equation}
\label{eq:final_augment_feat}
\mathbf{X}^{\text{aug}}\leftarrow\mathbf{X} * \mathbf{F}_s + \mathbf{X}^{\text{aug}} * \mathbf{C}, 
\end{equation}
which aims to maintain the consistency of augmented samples with original samples in factual regions, while allowing counterfactual regions to exhibit diverse distributions. 
Then we can re-generate the enhanced factual attention $\mathbf{F}^{\text{aug}}=\mathcal{F}(\mathbf{X}^{\text{aug}})$ and counterfactual attention $\mathbf{C}^{\text{aug}}=\mathcal{C}(\mathbf{X}^{\text{aug}})$ by re-using the networks in Section~\ref{sec:counterfact_attn_extraction}. 

To prevent the model from over-concentrating on single attention areas and to encourage the exploration of alternative forgery manifestations, we further introduce an attention diversification loss $\mathcal{L}_{{\text{aug}}}$ as follows: 
\begin{equation}
\mathcal{L}_{{\text{aug}}} = \frac{1}{M} \sum_{i=1}^{M} \left|\mathbf{X}*\mathbf{F}_i - {\mathbf{X}^{\text{aug}}*\mathbf{F}^{\text{aug}}_i}\right| \cdot (1 - \mathbf{1}_{i=s}),
\end{equation}
where $\mathbf{1}_{(\cdot)}$ is the indicator function. $M$ is the number of factual attention maps. 
Through minimizing $\mathcal{L}_{{\text{aug}}}$, it maintains causal consistency~\cite{yue2021counterfactual} between original and augmented features in model-specific regions while encouraging exploration of complementary forgery traces by excluding the already-attended regions from the consistency constraint.

The final loss function \( \mathcal{L}_{\text{total}} \) is computed as follows: 
\begin{equation}
\label{eq:total_loss}
\mathcal{L}_{\text{total}} = \eta_{1} \mathcal{L}_{\text{causal}} + \eta_{2} \mathcal{L}_{\text{decor}} + \eta_{3} \mathcal{L}_{{\text{aug}}} + \mathcal{L}_{\text{original}}
\end{equation}
where $\eta_{1}$, $\eta_{2}$ and $\eta_{3}$ are hyper-parameters. $\mathcal{L}_{\text{original}}$ is the original loss in baselines which also takes $\mathbf{X}$ as input.

\section{Experiments}
In this section, we evaluate \ours on the public benchmarks for open-world model attribution, including open-world deepfake attribution~\cite{sun2023contrastive}, GAN attribution and discovery~\cite{yang2023progressive}. 
Unlike closed-set attribution where source models of test samples are present in the training set, these benchmarks challenge the model to generalize to previously unseen generative models. 
The key challenge lies in preventing the learned features or patterns from overfitting to known models while maintaining the discrimination power. 
We incorporated \ours into state-of-the-art approaches in each benchmark (see Appendix for details), and conducted comprehensive comparisons with various baseline methods. 
In addition, we performed extensive ablation studies to analyze the effect of key components, hyper-parameters, design choices, as well as the model efficiency of our method. 

\begin{table*}[t]
\caption{Quantitative results (\%) of \textbf{Protocol 1} and \textbf{Protocol 2} on OW-DFA~\cite{sun2023contrastive}.}
\vspace{-7mm}
\label{tab:owdfa compare}
\begin{center}
\resizebox{.94\linewidth}{!}{%
\begin{tabular}{lcccccccccccccc}
\toprule
{\multirow{3}{*}{\textbf{Method}}} & \multicolumn{7}{c}{\textbf{Protocol-1: Fake}} & \multicolumn{7}{c}{\textbf{Protocol-2: Real \& Fake}} \\ \cmidrule(lr){2-8} \cmidrule(lr){9-15}
\multicolumn{1}{c}{} & \textbf{Known} & \multicolumn{3}{c}{\textbf{Novel}} & \multicolumn{3}{c}{\textbf{All}} & \textbf{Known} & \multicolumn{3}{c}{\textbf{Novel}} & \multicolumn{3}{c}{\textbf{All}} \\ \cmidrule(lr){2-2} \cmidrule(lr){3-5} \cmidrule(lr){6-8}  \cmidrule(lr){9-9} \cmidrule(lr){10-12} \cmidrule(lr){13-15} 
\multicolumn{1}{c}{} & \textbf{ACC} & \textbf{ACC} & \textbf{NMI} & \textbf{ARI} & \textbf{ACC} & \textbf{NMI} & \textbf{ARI} & \textbf{ACC} & \textbf{ACC} & \textbf{NMI} & \textbf{ARI} & \textbf{ACC} & \textbf{NMI} & \textbf{ARI} \\ \midrule

\textcolor{gray}{Lower Bound} & \textcolor{gray}{99.68} & \textcolor{gray}{40.86} & \textcolor{gray}{47.55} & \textcolor{gray}{26.33} & \textcolor{gray}{46.91} & \textcolor{gray}{63.43} & \textcolor{gray}{37.33} & \textcolor{gray}{99.84} & \textcolor{gray}{34.57} & \textcolor{gray}{42.98} & \textcolor{gray}{19.37} & \textcolor{gray}{61.46} & \textcolor{gray}{66.05} & \textcolor{gray}{62.16} \\
\textcolor{gray}{Upper Bound} & \textcolor{gray}{98.93} & \textcolor{gray}{96.99} & \textcolor{gray}{94.18} & \textcolor{gray}{94.94} & \textcolor{gray}{97.91} & \textcolor{gray}{95.87} & \textcolor{gray}{95.91} & \textcolor{gray}{99.27} & \textcolor{gray}{97.12} & \textcolor{gray}{94.89} & \textcolor{gray}{96.78} & \textcolor{gray}{98.43} & \textcolor{gray}{96.48} & \textcolor{gray}{98.27} \\ \hdashline[2pt/4pt]
Openworld-GAN~\cite{girish2021towards} & 99.49 & 39.14 & 47.08 & 44.39 & 57.92 & 58.24 & 48.71 & 99.51 & 40.09 & 50.72 & 35.96 & 67.02 & 57.81 & 59.92 \\
DNA-Det~\cite{yang2022deepfake}  & 99.62 & 38.16 & 48.76 & 23.21 & 46.88 & 66.15 & 35.46 & 99.45 & 39.03 & 47.07 & 22.54 & 60.68 & 67.94 & 56.49 \\
RankStats~\cite{han2020automatically} & 99.17 & 62.05 & 64.60 & 52.87 & 79.52 & 78.87 & 72.90 & 98.86 & 51.19 & 57.56 & 37.56 & 78.25 & 77.37 & 88.07 \\
OpenLDN~\cite{rizve2022openldn}   & 98.78 & 54.12 & 57.54 & 45.43 & 72.90 & 77.22 & 70.03 & 97.03 & 48.26 & 52.77 & 33.72 & 73.97 & 75.13 & 84.37 \\
ORCA~\cite{cao2022openworld} & 98.30 & 73.61 & 70.20 & 63.50 & 85.23 & 83.99 & 80.86 & 97.09 & 62.10 & 64.96 & 49.15 & 83.44 & 82.68 & 88.64 \\
NACH~\cite{guo2022robust} & 98.34 & 73.43 & 71.61 & 65.33 & 85.16 & 84.90 & 82.31 & 97.28 & 69.39 & 70.03 & 54.28 & 86.47 & 84.76 & 90.09 \\
CPL~\cite{sun2023contrastive} & 98.68 & 75.21 & 73.19 & 65.71 & 86.25 & 85.58 & 82.35 & 97.45 & 69.57 & 70.67 & 54.67 & 86.51 & 85.44 & 90.30 \\
MPSL~\cite{sun2024rethinking} & 98.55 & 75.23 & 76.99 & 69.26 & 86.31 & 87.27 & 84.91 & 98.01 & 69.81 & 72.77 & 55.97 & 86.71 & 86.60 & 92.37 \\
\midrule
\textbf{ORCA + Ours} & 98.55 & 76.85 & 77.18 & 67.77 & 86.91 & 88.04 & 83.74 & 98.44 & 66.05 & 70.65 & 50.42 & 86.20 & 86.59 & 91.65 \\
\rowcolor{cyan!10}
\textit{Improvement} & \textcolor{red}{+0.25} & \textcolor{red}{+3.24} & \textcolor{red}{+6.98} & \textcolor{red}{+4.27} & \textcolor{red}{+1.68} & \textcolor{red}{+4.05} & \textcolor{red}{+2.88} & \textcolor{red}{+1.35} & \textcolor{red}{+3.95} & \textcolor{red}{+5.69} & \textcolor{red}{+1.27} & \textcolor{red}{+2.76} & \textcolor{red}{+3.91} & \textcolor{red}{+3.01} \\ \hdashline[2pt/4pt]
\textbf{NACH + Ours} & 98.68 & 85.31 & 81.22 & 74.50 & 91.60 & 90.25 & 86.74 & 98.79 & 69.92 & 73.93 & 56.90 & 87.44 & 88.04 & 92.28 \\
\rowcolor{cyan!10}
\textit{Improvement} & \textcolor{red}{+0.34} & \textcolor{red}{+11.88} & \textcolor{red}{+9.61} & \textcolor{red}{+9.17} & \textcolor{red}{+6.44} & \textcolor{red}{+5.35} & \textcolor{red}{+4.43} & \textcolor{red}{+1.51} & \textcolor{red}{+0.53} & \textcolor{red}{+3.90} & \textcolor{red}{+2.62} & \textcolor{red}{+0.97} & \textcolor{red}{+3.28} & \textcolor{red}{+2.19} \\ \hdashline[2pt/4pt]
\textbf{CPL + Ours} & \textbf{98.90} & \textbf{86.02} & \textbf{82.19} & \textbf{76.98} & \textbf{92.06} & \textbf{90.60} & \textbf{87.66} & \textbf{99.08} & \textbf{70.56} & \textbf{74.14} & \textbf{57.22} & \textbf{88.02} & \textbf{88.59} & \textbf{93.01} \\
\rowcolor{cyan!10}
\textit{Improvement} & \textcolor{red}{+0.22} & \textcolor{red}{+10.81} & \textcolor{red}{+9.00} & \textcolor{red}{+11.27} & \textcolor{red}{+5.81} & \textcolor{red}{+5.02} & \textcolor{red}{+5.31} & \textcolor{red}{+1.63} & \textcolor{red}{+0.99} & \textcolor{red}{+3.47} & \textcolor{red}{+2.55} & \textcolor{red}{+1.51} & \textcolor{red}{+3.15} & \textcolor{red}{+2.71} \\ \bottomrule
\end{tabular}%
}
\vspace{-4mm}
\end{center}
\end{table*}

\begin{figure*}[t]
    \centering
    \includegraphics[width=.95\textwidth]{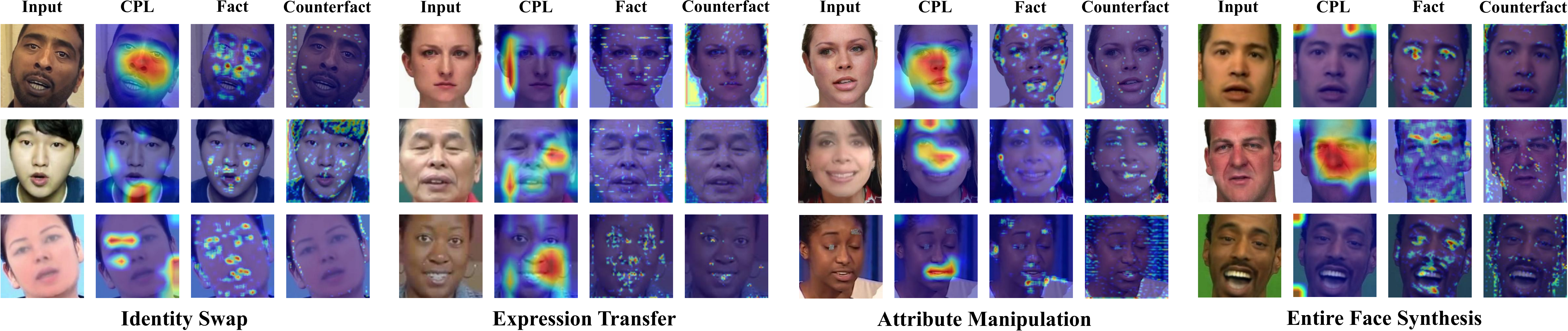}
    \vspace{-3.5mm}
    \caption{Visualization of learned attention maps across various forgery types. }
    \vspace{-4mm}
    \label{fig:cam}
\end{figure*}

\subsection{Open-world Deepfake Attribution}
\noindent\textbf{Dataset: }The task aims to simultaneously attribute fake face images to known source models and identify those from unknown ones. 
We experimented on the OW-DFA dataset~\cite{sun2023contrastive} which contains 20 challenging forgery methods and real face images from FaceForensics++~\cite{rossler2019ff++} and Celeb-DF ~\cite{li2020celeb}, whose forgery types include identity swap, expression transfer, attribute manipulation, entire face synthesis. 
Each type (including real faces) contains both labeled and unlabeled images. 

\noindent\textbf{Protocols: }We followed \cite{sun2023contrastive} to employ two protocols. 
Protocol 1 focuses on forged images only, where labeled and unlabeled images were treated as known and novel attacks encountered in the open world. 
In Protocol 2, real images from FaceForensics++~\cite{rossler2019ff++} (labeled) and Celeb-DF ~\cite{li2020celeb} (unlabeled) are blended with forged images to create a dataset where real faces substantially outnumber fake ones as in real-world conditions. 

\noindent\textbf{Evaluation Metrics: } We followed ~\cite{sun2023contrastive} to use classification Accuracy (ACC), Normalized Mutual Information (NMI), and Adjusted Rand Index (ARI) to evaluate on the OW-DFA dataset, where Hungarian algorithm~\cite{kuhn1955hungarian} was applied to align predictions with ground-truth labels.

\noindent\textbf{Results: }To get deeper insights into \ours, we firstly visualize the learned attentions via CAM~\cite{zhou2016learning} in Figure~\ref{fig:cam}. 
While CPL~\cite{sun2023contrastive} attends to broader, less focused regions that potentially dilute discriminative traces, our factual attention identifies model-specific patterns critical for attribution. 
Meanwhile, the counterfactual attention highlights source bias regions (including irrelevant background) that would otherwise introduce spurious correlations and confound the attribution process.  

The quantitative results on OW-DFA are shown in Table~\ref{tab:owdfa compare}, where the ``upper bound'' and ``lower bound'' represent supervised learning on the
labeled set and labeled+unlabeled sets respectively~\cite{sun2023contrastive}. 
In Protocol 1, our method demonstrates significant advantages in open-world scenarios. 
Particularly, \ours significantly boosts the novel attack attribution by 11.27\% ARI for CPL~\cite{sun2023contrastive}, which outperforms previous state-of-the-art MPSL~\cite{sun2024rethinking} by 7.72\%. 
The improvements upon ORCA and NACH, while not designed for model attribution, collectively validate the generalization ability of \ours. 
In Protocol-2 which involves real faces simulating challenging real-world scenarios, our method also contributes to strong performance gains. 
Compared to MPSL~\cite{sun2024rethinking}, CPL~\cite{sun2023contrastive}+Ours leads by 1.37\% in NMI for novel attacks. 
Integration into other baselines like ORCA~\cite{cao2022openworld} also improves the key metrics, which validates our proposed method in real-world scenarios. 

\begin{figure}[t]
    \centering
    \includegraphics[width=.9\linewidth]{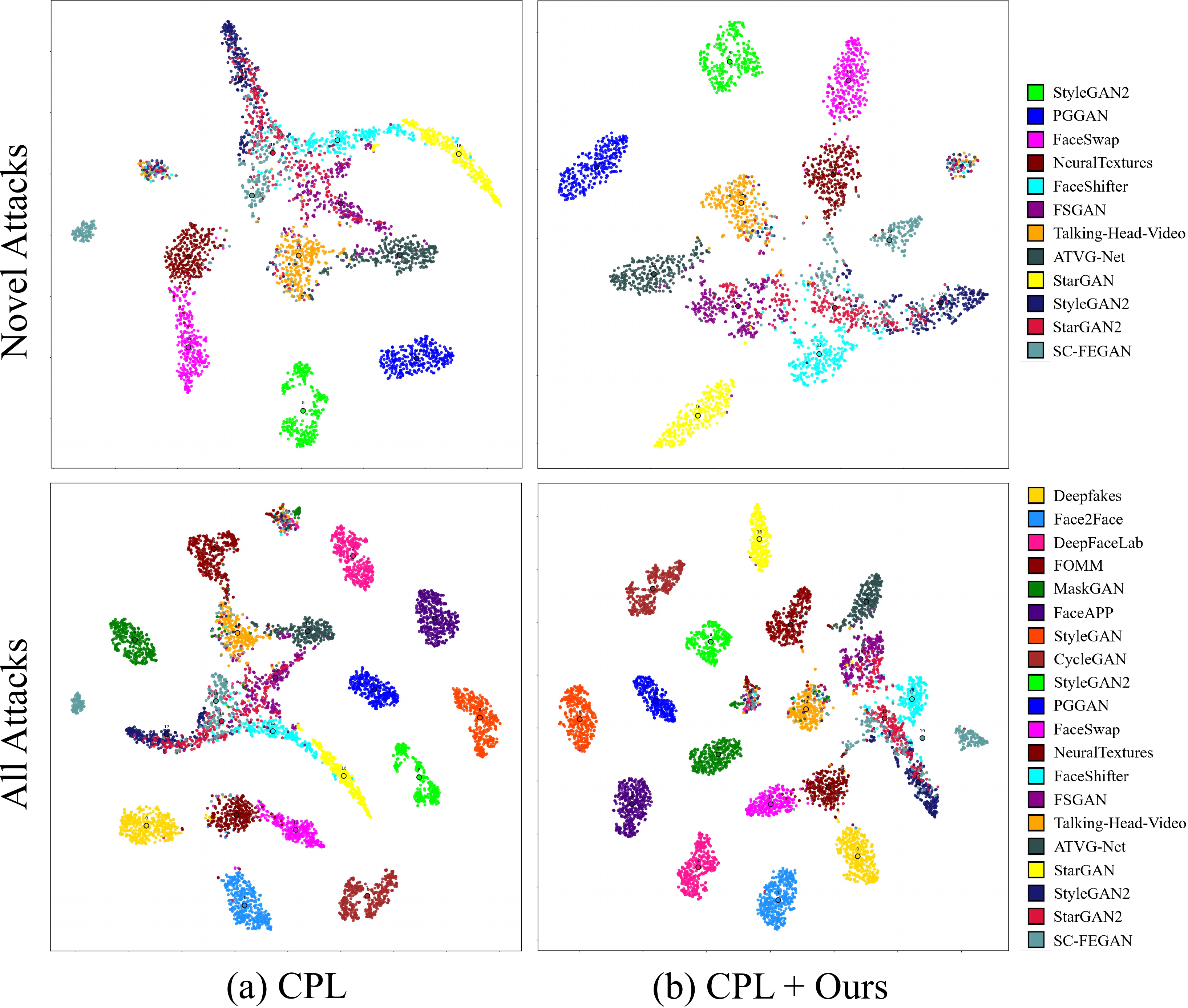}
    \vspace{-3mm}
    \caption{t-sne visualization in OW-DFA. }
    \vspace{-4mm}
    \label{fig:tsne_owdfa}
\end{figure}

\noindent\textbf{t-SNE Visualization: }To visually compare CDAL and baseline methods, we presented t-SNE ~\cite{van2008visualizing} results in Figure~\ref{fig:tsne_owdfa} using the features for final prediction. 
The visualization includes 8 known attacks and 12 novel attacks. 
\ours significantly improves the clustering performance on novel attacks while maintaining discriminative power for known attacks. 
For novel attacks~\cite{choi2018stargan,li2019faceshifter,Jo_2019_ICCV,nirkin2019fsgan}, CDAL exhibits effective dicrimination ability. 
Even in scenarios involving cross-source datasets (e.g., DeepFakes ~\cite{deepfakes} from FaceForensics++~\cite{rossler2019ff++} and DeepFaceLab~\cite{DeepFaceLab} from ForgeryNet~\cite{he2021forgerynet} both belonging to identity swap; PGGAN~\cite{karras2018progressive} from DFFD~\cite{dang2020detection} and CycleGAN~\cite{CycleGAN2017} from ForgeryNIR~\cite{wang2022forgerynir} both belonging to Entire Face Synthesis), CDAL still substantially reduces the feature distances of the samples from the same forgery type. 
See Appendix for the t-SNE visualizations under Protocol 2. 

\begin{table}[t]
\centering
\caption{Results (\%) of \textbf{Protocol 1} on OW-DFA~\cite{sun2023contrastive} extended with modern generative models from DF40~\cite{yan2024df40}.}
\vspace{-4mm}
\label{tab:generalization}
\resizebox{.97\columnwidth}{!}{%
\begin{tabular}{llccccccc}
\toprule
{\multirow{2}{*}{\textbf{Setting}}} & {\multirow{2}{*}{\textbf{Method}}} & \textbf{Known} & \multicolumn{3}{c}{\textbf{Novel}} & \multicolumn{3}{c}{\textbf{All}} \\ \cmidrule(lr){3-3} \cmidrule(lr){4-6} \cmidrule(lr){7-9}
\multicolumn{2}{c}{} & \textbf{ACC} & \textbf{ACC} & \textbf{NMI} & \textbf{ARI} & \textbf{ACC} & \textbf{NMI} & \textbf{ARI} \\ \midrule
 & CPL~\cite{sun2023contrastive} & 98.83 & 76.40 & 78.16 & 65.96 & 82.94 & 87.50 & 78.76 \\
\textbf{Setting1} & CPL + Ours & 99.80 & 83.07 & 82.73 & 74.17 & 86.75 & 89.75 & 82.52 \\
 & \cellcolor{cyan!10} \textit{Improvement} & \cellcolor{cyan!10} \textcolor{red}{+0.97} & \cellcolor{cyan!10} \textcolor{red}{+6.67} & \cellcolor{cyan!10} \textcolor{red}{+4.57} & \cellcolor{cyan!10} \textcolor{red}{+8.21} & \cellcolor{cyan!10} \textcolor{red}{+3.81} & \cellcolor{cyan!10} \textcolor{red}{+2.25} & \cellcolor{cyan!10} \textcolor{red}{+3.76} \\ \midrule
 & CPL~\cite{sun2023contrastive} & 99.21 & 76.31 & 74.04 & 66.45 & 87.10 & 86.87 & 83.28 \\
\textbf{Setting2} & CPL + Ours & 99.34 & 80.00 & 79.53 & 72.64 & 89.07 & 89.30 & 85.90 \\
 & \cellcolor{cyan!10} \textit{Improvement} & \cellcolor{cyan!10} \textcolor{red}{+0.13} & \cellcolor{cyan!10} \textcolor{red}{+3.69} & \cellcolor{cyan!10} \textcolor{red}{+5.49} & \cellcolor{cyan!10} \textcolor{red}{+6.19} & \cellcolor{cyan!10} \textcolor{red}{+1.97} & \cellcolor{cyan!10} \textcolor{red}{+2.43} & \cellcolor{cyan!10} \textcolor{red}{+2.62} \\

\bottomrule
\end{tabular}%
}
\vspace{-5mm}
\end{table}

\begin{table*}[t]
\setlength{\abovecaptionskip}{2mm}
\caption{Quantitative results (\%) of GAN attribution (Protocol 1) on OSMA~\cite{yang2023progressive}, which are averaged among five splits.}
\vspace{-5mm}
\begin{center}
\resizebox{.85\linewidth}{!}{%
\begin{tabular}{lccccccccccccc}
\toprule
\multirow{2}{*}{\textbf{Method}} & \textbf{Closed-Set} & & \multicolumn{2}{c}{\textbf{Unseen Seed}} & & \multicolumn{2}{c}{\textbf{Unseen Architecture}} & & \multicolumn{2}{c}{\textbf{Unseen Dataset}} & & \multicolumn{2}{c}{\textbf{Unseen All}} \\  
\cmidrule{2-2} \cmidrule{4-5} \cmidrule{7-8} \cmidrule{10-11}  \cmidrule{13-14}  
 & \textbf{Accuracy} & & \textbf{AUC}  & \textbf{OSCR}  & &  \textbf{AUC}  & \textbf{OSCR} & & \textbf{AUC} &  \textbf{OSCR} & &  \textbf{AUC}  & \textbf{OSCR} \\
\midrule
PRNU~\cite{marra2019gans} & 55.27 && 69.20 & 49.16 && 70.02 & 49.49 && 67.68 & 48.57 && 68.94 & 49.06 \\
Yu~et al~\cite{yu2019attributing} & 85.71 && 53.14 & 50.99 && 69.04 & 64.17 && 78.79 & 72.20 && 69.90 & 64.86  \\
DCT-CNN~\cite{frank2020leveraging} & 86.16 && 55.46 & 52.68 && 72.56 & 67.43 && 72.87 & 67.57 && 69.46 & 64.70  \\
DNA-Det~\cite{yang2022deepfake}  & 93.56 && 61.46 & 59.34 && 80.93 & 76.45 && 66.14 & 63.27 && 71.40 & 68.00 \\
RepMix~\cite{bui2022repmix} & 93.69 && 54.70 & 53.26 && 72.86 & 70.49 && 78.69 & 76.02 && 71.74 & 69.43 \\
POSE~\cite{yang2023progressive} & 94.81 && 68.15 & 67.25 && 84.17 & 81.62 && 88.24 & 85.64 && 82.76 & 80.50  \\
\midrule
\textbf{RepMix + Ours} & 94.01 && 59.20 & 57.41 && 75.38 & 73.11 && 81.73 & 78.96 && 73.97 & 71.72 \\
\rowcolor{cyan!10}
\textit{Improvement} & \textcolor{red}{+0.32} && \textcolor{red}{+4.50} & \textcolor{red}{+4.15} && \textcolor{red}{+2.52} & \textcolor{red}{+2.62} && \textcolor{red}{+3.04} & \textcolor{red}{+2.94} && \textcolor{red}{+2.23} & \textcolor{red}{+2.29} \\
\hdashline[2pt/4pt]
\textbf{POSE + Ours} & \textbf{95.25} && \textbf{72.99} & \textbf{71.73} && \textbf{86.97} & \textbf{84.48} && \textbf{91.32} & \textbf{88.66} && \textbf{85.37} & \textbf{82.85} \\
\rowcolor{cyan!10}
\textit{Improvement} & \textcolor{red}{+0.44} && \textcolor{red}{+4.84} & \textcolor{red}{+4.48} && \textcolor{red}{+2.80} & \textcolor{red}{+2.86} && \textcolor{red}{+3.08} & \textcolor{red}{+3.02} && \textcolor{red}{+2.61} & \textcolor{red}{+2.35} \\
\bottomrule
\end{tabular}%
}
\label{tab:gan_attribution}
\vspace{-7mm}
\end{center}
\end{table*}

\noindent\textbf{Generalization to Modern Generative Models: }
To keep pace with the development tendency, we extended the OW-DFA benchmark with diffusion-based and flow-based models (see Appendix for details). 
We design two experimental settings: In Setting 1, we augment both training and test sets with diffusion models~\cite{peebles2023scalable,atito2021sit,liu2024residual} and flow-based model~\cite{esser2021taming}) 
from the recent large-scale DF-40 dataset~\cite{yan2024df40}, where DiT-XL/2~\cite{peebles2023scalable} serves as a labeled known attack and the others as unlabeled novel attacks. 
As shown in Table~\ref{tab:generalization}, our method achieved substantial performance gains on novel attacks. 
In Setting 2, we evaluate the generalization on the completely unseen DDPM model~\cite{ho2020denoising}. 
Our method demonstrates superior performance compared to CPL~\cite{sun2023contrastive}. 
This is because the CPL model learns the distribution within the training set, thus struggling with out-of-distribution attacks in this open-set scenario. 
These results validate our approach at both identifying known manipulation sources and discovering emerging generation techniques in real-world applications. 

\subsection{Open-world GAN Attribution}
\noindent\textbf{Dataset: }This task aims to simultaneously attribute GAN-generated images to known GAN models and discover novel GAN classes in an open-world scenario. 
We experimented on the OSMA~\cite{yang2023progressive} benchmark with five splits, which contains 15 known classes as the close-set (including real and 14 seen models), 
and 53 unknown classes as the open-set in total. 
The 53 unknown classes comprise of models trained with 10 random seeds, 22 architecture and 21 dataset distribtions respectively. 

\noindent\textbf{Protocols: }
We followed the protocols from \cite{yang2023progressive}. 
In Protocol 1 (GAN Attribution), models were trained on the closed-set, and attribution confidence scores were computed for both closed-set and open-set using standard OSCR. 
In Protocol 2 (GAN Discovery), models were trained on known GAN classes. During testing, features of closed-set and open-set images were extracted and clustered via K-Means. 
Closed-set samples were mapped to known GAN classes, while open-set samples formed auxiliary clusters represented as novel classes. 

\begin{table}[t]
\setlength{\abovecaptionskip}{2mm}
\caption{Quantitative results (\%) of GAN discovery (Protocol 2) on OSMA~\cite{yang2023progressive}, which are averaged among five splits.} 
\vspace{-5mm}
\begin{center}
\resizebox{.84\linewidth}{!}{%
\begin{tabular}{lcccc}
\toprule
{\multirow{2}{*}{\textbf{Method}}} & \multicolumn{1}{c}{\textbf{Close-set}} & \multicolumn{3}{c}{\textbf{Unseen All}} \\  
\cmidrule(lr){2-2} \cmidrule(lr){3-5}
 & \textbf{ACC} & \textbf{Purity} & \textbf{NMI} & \textbf{ARI} \\
\midrule
RepMix~\cite{bui2022repmix} & 94.01 & 31.53 & 51.60 & 18.71 \\
POSE~\cite{yang2023progressive} & 94.81 & 41.04 & 60.59 & 26.39 \\
\midrule
\textbf{RepMix + Ours} & 93.83 & 37.96 & 52.08 & 20.66 \\
\rowcolor{cyan!10}
\textit{Improvement} & \textcolor{red}{+0.32} & \textcolor{red}{+6.43} & \textcolor{red}{+0.48} & \textcolor{red}{+1.95} \\
\hdashline[2pt/4pt]
\textbf{POSE + Ours} & \textbf{95.25} & \textbf{48.93} & \textbf{61.89} & \textbf{29.65} \\
\rowcolor{cyan!10}
\textit{Improvement} & \textcolor{red}{+0.44} & \textcolor{red}{+7.89} & \textcolor{red}{+1.30} & \textcolor{red}{+3.26} \\
\bottomrule
\end{tabular}%
}
\label{tab:gan_discovery}
\vspace{-8mm}
\end{center}
\end{table}

\noindent\textbf{Evaluation Metrics: }Following ~\cite{girish2021towards, yang2023progressive}, for Protocol 1, we reported Area Under the ROC Curve (AUC) and Open Set Classification Rate (OSCR)~\cite{dhamija2018reducing}, which balances the accurate classification of known models, and the correct discrimination between known and unknown models. 
For Protocol 2, we evaluated closed-set performance using Accuracy (ACC) and open-set performance using Average Purity, NMI, and ARI. Results were averaged on five splits~\cite{yang2023progressive}. 

\begin{table*}[t]
\centering
\caption{Results of ablation studies and in-depth analysis of \ours.}
\vspace{-3mm}
\label{tab:all_tables}
\begin{subtable}[t]{0.37\textwidth}
\centering
\caption{Ablation study on key components of \ours.}
\vspace{-1.5mm}
\label{tab:ablation_all}
\resizebox{\columnwidth}{!}{%
\setlength{\tabcolsep}{2.5pt} 
\begin{tabular}{cccccccccc}
\toprule
\multicolumn{1}{c}{\multirow{2}{*}{\textbf{Baseline}}} &
  \multicolumn{1}{c}{\multirow{2}{*}{\textbf{FA}}} &
  \multicolumn{1}{c}{\multirow{2}{*}{\textbf{CA}}} &
  \multicolumn{1}{c}{\multirow{2}{*}{\textbf{EA}}} &
  \multicolumn{3}{c}{\textbf{Novel (OW-DFA)}}  &
  \multicolumn{3}{c}{\textbf{Unseen All (OSMA)}} \\ \cmidrule(lr){5-7} \cmidrule(l){8-10} 
\multicolumn{1}{c}{} &
  \multicolumn{1}{c}{} &
  \multicolumn{1}{c}{} &
  \multicolumn{1}{c}{} &
  \textbf{ACC} &
  \textbf{NMI} &
  \textbf{ARI} &
  \textbf{Purity} &
  \textbf{NMI} &
  \textbf{ARI} \\ \midrule
  
\checkmark &                           &                           &                           & 75.21 & 73.19 & 65.71 & 41.04 & 60.59 & 26.39 \\
\checkmark & \checkmark &                           &                           & 81.78 & 78.20 & 70.34 & 45.58 & 60.96 & 27.63 \\
\checkmark & \checkmark & \checkmark &                           & 84.05 & 80.86 & 73.54 & 46.65 & 61.07 & 27.95 \\ 
\rowcolor{cyan!10} \checkmark & \checkmark & \checkmark & \checkmark & \textbf{86.02} & \textbf{82.19} & \textbf{76.98} & \textbf{48.93} & \textbf{61.89} & \textbf{29.65} \\ \bottomrule
\end{tabular}%
}
\end{subtable}
\hfill
\hspace{-2mm}
\begin{subtable}[t]{0.415\textwidth}
\centering
\caption{Ablation study on loss functions of \ours.}
\vspace{-1.5mm}
\label{tab:ablation_loss}
\setlength{\tabcolsep}{2.5pt} 
\resizebox{\columnwidth}{!}{%
\begin{tabular}{ccccccccccc}
\toprule
\multicolumn{1}{c}{\multirow{2}{*}{\textbf{Baseline}}} &
  \multicolumn{1}{c}{\multirow{2}{*}{$\mathbf{\mathcal{L}_{\text{causal}}}$}} &
  \multicolumn{1}{c}{\multirow{2}{*}{$\mathbf{\mathcal{L}_{\text{decor}}}$}} &
  \multicolumn{1}{c}{\multirow{2}{*}{$\mathbf{\mathcal{L}_{\text{aug}}}$}} &
  \multicolumn{3}{c}{\textbf{Novel (OW-DFA)}}  &
  \multicolumn{3}{c}{\textbf{Unseen All (OSMA)}} \\ \cmidrule(lr){5-7} \cmidrule(l){8-10} 
\multicolumn{1}{c}{} &
  \multicolumn{1}{c}{} &
  \multicolumn{1}{c}{} &
  \multicolumn{1}{c}{} &
  \textbf{ACC} &
  \textbf{NMI} &
  \textbf{ARI} &
  \textbf{Purity} &
  \textbf{NMI} &
  \textbf{ARI} \\ \midrule
  
\checkmark &                           &                           &                           & 79.56 & 77.24 & 68.94 & 43.67 & 60.86 & 27.08 \\
\checkmark & \checkmark &                           &                           & 82.46 & 79.90 & 71.75 & 46.23 & 61.02 & 27.91 \\
\checkmark & \checkmark & \checkmark &                           & 84.85 & 81.80 & 75.08 & 47.22 & 61.34 & 28.19 \\ 
\rowcolor{cyan!10} \checkmark & \checkmark & \checkmark & \checkmark & \textbf{86.02} & \textbf{82.19} & \textbf{76.98} & \textbf{48.93} & \textbf{61.89} & \textbf{29.65} \\ \bottomrule
\end{tabular}%
}
\end{subtable}%
\hfill
\begin{subtable}[t]{0.173\textwidth}
\centering
\caption{Ablation study on N.}
\vspace{-1.5mm}
\label{tab:expert_num}
\resizebox{\columnwidth}{!}{%
\renewcommand{\arraystretch}{0.89}
\setlength{\tabcolsep}{2pt} 
\begin{tabular}{lcccc}
\toprule
{\multirow{2}{*}{\textbf{N}}} & \textbf{Known} & \multicolumn{3}{c}{\textbf{Novel}} \\ \cmidrule(lr){2-2}  \cmidrule(lr){3-5} 
\multicolumn{1}{c}{} & \textbf{ACC} & \textbf{ACC} & \textbf{NMI} & \textbf{ARI} \\ \midrule
2 & 98.76 & 84.27 & 79.23 & 72.57 \\
3 & 98.25 & 82.10 & 80.52 & 72.60 \\
\rowcolor{cyan!10} 4 & \textbf{98.90} & \textbf{86.02} & \textbf{82.19} & \textbf{76.98} \\
5 & 98.68 & 82.68 & 81.94 & 74.56 \\
6 & 98.68 & 79.77 & 79.94 & 72.00 \\
\bottomrule
\end{tabular}%
}
\end{subtable}

\vspace{2mm}

\begin{subtable}[t]{0.31\textwidth}
\centering
\vspace{-2mm}
\caption{Comparison on counterfactual attentions.}
\vspace{-1.5mm}
\label{tab:counterfact}
\setlength{\tabcolsep}{2.5pt} 
\resizebox{\columnwidth}{!}{%
\begin{tabular}{lcccc}
\toprule
{\multirow{2}{*}{\textbf{Method}}} & \textbf{Known} & \multicolumn{3}{c}{\textbf{Novel}} \\ \cmidrule(lr){2-2}  \cmidrule(lr){3-5} 
\multicolumn{1}{c}{} & \textbf{ACC} & \textbf{ACC} & \textbf{NMI} & \textbf{ARI} \\ \midrule
CPL~\cite{sun2023contrastive} & 98.68 & 75.21 & 73.19 & 65.71 \\ \hdashline[2pt/4pt]
CPL + Ours (Random) & 98.35 & 84.05 & 80.86 & 73.54 \\
CPL + Ours (Uniform) & 98.68 & 82.68 & 79.52 & 70.81 \\
CPL + Ours (Reversed) & 98.46 & 82.30 & 79.27 & 71.88 \\
CPL + Ours (Shuffle) & 98.46 & 80.74 & 79.56 & 72.12 \\ \midrule
\rowcolor{cyan!10} CPL + Ours (CE-Conv) & \textbf{98.90} & \textbf{86.02} & \textbf{82.19} & \textbf{76.98} \\ \bottomrule
\end{tabular}%
}
\end{subtable}%
\hfill
\begin{subtable}[t]{0.307\textwidth}
\centering
\vspace{-2mm}
\caption{Comparison with vanilla attention.}
\vspace{-1.5mm}
\label{tab:attention}
\setlength{\tabcolsep}{2pt} 
\renewcommand{\arraystretch}{0.95}
\resizebox{\columnwidth}{!}{%
\begin{tabular}{lccccc}
\toprule
{\multirow{2}{*}{\textbf{Method}}} & \textbf{Known} & \multicolumn{3}{c}{\textbf{Novel}} \\ \cmidrule(lr){2-2}  \cmidrule(lr){3-5} 
 & \textbf{ACC} & \textbf{ACC} & \textbf{NMI} & \textbf{ARI} \\ \midrule
CPL~\cite{sun2023contrastive} & 98.68 & 75.21 & 73.19 & 65.71 \\ \midrule
CPL + Attention & 98.25 & 70.37 & 70.59 & 58.81 \\
\rowcolor{gray!20}
$\Delta$& {\color{blue}{-0.43}} & {\color{blue}{-4.84}} & {\color{blue}{-2.60}} & {\color{blue}{-6.90}} \\ \midrule

CPL + \textbf{Ours} & \textbf{98.90} & \textbf{86.02} & \textbf{82.19} & \textbf{76.98} \\
\rowcolor{cyan!10}
$\Delta$& {\color{red}{+0.22}} & {\color{red}{+10.81}} & {\color{red}{+9.00}} & {\color{red}{+11.27}} \\ \bottomrule
\end{tabular}%
}
\end{subtable}%
\hfill
\begin{subtable}[t]{0.275\textwidth}
\centering
\vspace{-2mm}
\caption{Comparison on model efficiency.}
\vspace{-1.5mm}
\label{tab:computation}
\setlength{\tabcolsep}{3pt}
\renewcommand{\arraystretch}{0.75}
\resizebox{\columnwidth}{!}{%
\begin{tabular}{@{}lccc@{}}
\toprule
\multicolumn{4}{>{\columncolor[gray]{0.85}}c}{\textbf{OW-DFA}} \\
\midrule
{\textbf{Method}} & \textbf{ARI} & \textbf{Params/M} & \textbf{FLOPs/G} \\
\cmidrule(r){1-1} \cmidrule(r){2-2} \cmidrule(r){3-3} \cmidrule(l){4-4}
CPL~\cite{sun2023contrastive} & 65.71 & 23.59 & 5.397 \\
\rowcolor{cyan!10}CPL + Ours & 76.98 & 23.91 & 5.399 \\
\addlinespace[3pt]
\toprule
\multicolumn{4}{>{\columncolor[gray]{0.85}}c}{\textbf{OSMA}} \\
\midrule
{\textbf{Method}} & \textbf{ARI} & \textbf{Params/M} & \textbf{FLOPs/G} \\
\cmidrule(r){1-1} \cmidrule(r){2-2} \cmidrule(r){3-3} \cmidrule(l){4-4}
POSE~\cite{yang2023progressive} & 26.39 & 22.68 & 1.039 \\
\rowcolor{cyan!10}POSE + Ours & 29.65 & 22.93 & 1.041 \\
\bottomrule
\end{tabular}%
}
\end{subtable}
\vspace{-4mm}
\end{table*}

\noindent\textbf{Results: }
Results of Protocol 1 on OSMA are shown in Table~\ref{tab:gan_attribution}.  
\ours achieves the greatest improvements on the unseen seed setting, which indicates that \ours can still effectively capture subtle differences due to randomness. 
Table~\ref{tab:gan_discovery} presents the comparison on Protocol 2 of GAN discovery. 
Our method improves significantly on both RepMix~\cite{bui2022repmix} and POSE~\cite{yang2023progressive}, with overall purity in the unseen setting increased by 6.43 and 7.89 \% respectively. 
These results highlight the effectiveness of \ours in improving generalization to unseen GAN models across various scenarios. 
For qualitative results, please refer to the Appendix pages for the t-SNE visualization on the comparison of with and without our \ours.

\subsection{Ablation Studies and Analysis}
In this subsection, we present ablation studies to verify our key components, hyper-parameters and model efficiency. 
See Appendix for further results and analysis. 

\noindent\textbf{Ablation Study on Key Components: }Results on ablating the key components in \ours is shown in Table~\ref{tab:ablation_all}. 
Here, FA, CA, EA stand for factual attentions,  counterfactual attentions, and the enhanced attentions after being augmented. 
When only FA is used, the model dynamically extracts the factual attention, which significantly improves performance on OW-DFA and OSMA. 
However, the limited improvement for unseen classes is attributed to the static random counterfactual attention. 
After introducing CA, the NMI for novel classes in OW-DFA and unseen classes in OSMA increases by 2.66\% and 2.86\%, respectively. This indicates a stronger adaptability to unseen classes.
Finally, the introduction of EA achieves the best overall performance. 
EA, based on causal consistency, integrates factual and counterfactual attention to generate more robust feature representations.
These results collectively validate the effectiveness of the designed component in \ours.

\noindent\textbf{Ablation Study on Loss Functions of \ours: }  
Table~\ref{tab:ablation_loss} shows the ablation results on loss functions. 
Compared to baseline with $\mathcal{L}_{\text{original}}$ only, 
$\mathcal{L}_{\text{causal}}$ achieves an improvement of +2.90\% ACC on OW-DFA by decoupling factual attentions from counterfactual ones, which validates our overall causal modeling. 
This is further enhanced by $\mathcal{L}_{\text{decor}}$
which enables counterfactual attention to focus on source biases that might be shared across attribution models. 
$\mathcal{L}_{\text{aug}}$ further masks the already attended regions to encourage the model to explore complementary feature regions, which leads to a notable gain of +1.9\% ARI on OW-DFA. 
These ablational experiments collectively demonstrate the efficacy of the designed loss functions in our \ours.

\noindent\textbf{Ablation Study on the Number of Experts: }Table~\ref{tab:expert_num} shows the results on OW-DFA Protocol-1 with varying numbers of experts (N) in CE Convolution. 
\ours achieves the best performances across all metrics when $N$=4, which outperforms all the other trials by sizable margins.

\noindent\textbf{Ablation Study on Type of Counterfactual Attention: }Table~\ref{tab:counterfact} presents the results of different counterfactual attention types (see Appendix for definitions). Our strategy significantly outperforms both the baseline and static alternatives like random and uniform attention. 
This is because CE-Conv actively learns to extract adaptive counterfactual patterns, rather than simply applying fixed interventions that cannot effectively address the diverse characteristics of different forgery methods in open-word scenarios. 

\noindent\textbf{Can Vanilla Attention Promote Attribution Performances? }
It is crucial to validate that it is our proposed design in \ours that indeed promotes the performances, rather than the vanilla attention itself. 
We accordingly conducted experiments by removing \ours and only adding the vanilla attention to baselines. 
From Table~\ref{tab:attention}, we can see that adding vanilla attention to CPL leads to performance declines across all metrics, as apposed to the significant improvements by \ours. 
This degradation can be attributed to vanilla attention focusing excessively on sample-specific features, which introduces bias and weakends generalization. 
On the contrary, \ours effectively filters out these biases by focusing on model-specific artifacts, leading to more robust and generalizable attribution ability. 

\noindent\textbf{Computational Overhead Analysis: }Table~\ref{tab:computation} shows the computational overhead of our \ours when incorporated into the baselines. 
With significant performance improvements, our method only brings with up to 0.35M additional network parameters and 0.002 GFLOPs, which demonstrates the high efficiency of our method. 

\section{Conclusion}
In this paper, we propose Counterfactually Decoupled Attention Learning (\ours) for open-world model attribution. Unlike existing methods that rely on handcrafted strategies susceptible to spurious correlations, \ours explicitly models causal relationships between visual forgery traces and source models, which effectively decouples model-specific artifacts from source biases. 
By maximizing the causal effect between factual and counterfactual attention maps, our approach encourages networks to capture generalizable generation patterns. Experiments show \ours consistently improves state-of-the-art models with minimal overhead, especially for unseen attacks. Future work could explore extending this framework to broader forensics tasks such as video and multi-modality attributions. 

{
    \small
    \bibliographystyle{ieeenat_fullname}
    \bibliography{ref}
}


\end{document}